# Fusing Image Representations for Classification Using Support Vector Machines

Can Demirkesen
Laboratoire Jean Kuntzmann
Université Joseph Fourrier, LJK - UJF - BP 53 38041
Grenoble, France
candemirkesen@gmail.com

Hocine Cherifi
Institue of Science and Engineering
Galatasaray University, Istanbul,Turkey
hocine.cherifi@u-bourgogne.fr

*Abstract*— **In order to improve classification accuracy different image representations are usually combined. This can be done by using two different fusing schemes. In feature level fusion schemes, image representations are combined before the classification process. In classifier fusion, the decisions taken separately based on individual representations are fused to make a decision. In this paper the main methods derived for both strategies are evaluated. Our experimental results show that classifier fusion performs better. Specifically Bayes belief integration is the best performing strategy for image classification task.**

## I. INTRODUCTION

Semantic categorization of images requires analysis of many characteristics of an image such as color, texture and edge properties. Categorization based on these characteristics in a separate manner leads to a limited performance. Therefore it seems natural to combine all the available information in order to improve the performance in this task. Fusion of information that allows the needed improvement can be carried out on two levels of abstraction: (1) *feature level fusion*, (2) *classifier fusion*. Feature level fusion that is also known as "early fusion" or "pre-classification fusion" consists of integrating different types of information that represents an image. The integration process is performed before any classification or matching stage. Classifier fusion, on the other hand, consists of fusing classifier outputs. In this type of fusion, classification is performed on individual characteristics (features) and classification decisions are fused afterwards. Classifier fusion is also referred to as "late fusion", "decision fusion" or "mixture of experts" [1].

Processing of different types of information by human visual system has been studied by numerous experiments on human participants [2]. These studies have shown that different visual cues are processed separately by human brain then integrated together. This shows the important role of fusing visual cues in image classification task for human. There is no reason for not doing the same for automatic image classification however we do not know if the human visual system performs a feature or a classifier fusion. In order to determine which one is more appropriate for computer vision one should compare the classification performances for each of the two approaches. To our best knowledge there is no work in the literature of image classification that addresses this issue. All related work is based on only one approach. In [3], the authors address the fusion issue by focusing only on feature level fusion methods for the categorization of beach and urban images. More specifically, they use Support Vector Machines (SVM) with merging feature fusion. The authors propose to use PCA for merging the feature vectors if features do not have equivalent magnitude of numerical values. In a video content indexing task [4], the authors use two feature fusion strategies to combine features. The first strategy is the so called static feature fusion which consists of merging separate feature vectors into a unique vector. The second strategy is to merge feature vectors using Principal Component Analysis (PCA) to reduce feature dimensionality. The authors conclude that PCA obtain superior results to those obtained by the static feature fusion. These contributions show that PCA is beneficial when features are concatenated into single vector. In [5], a classifier technique is used to categorize rock images. The authors propose to use Hamming distance to make the final decision on the classification result vectors (CRVs). The CRVs are generated by separate individual classifiers based on different features. In this work the results cover only classifier fusion and rock images, not natural scenes. In [6], a content based image retrieval task is carried out with a fusion based weighted similarity matching function with experimentally selected weights. The highest weight is assigned to the similarity measure that appears to be the most accurate among the global, semi-global and local similarity measures. The proposed technique and results are not compared with another approach.
SVM classifiers trained on color and texture are fused for video classification in [7]. A technique called Transferable Belief Model based on Dempster-Shafer theory is used for fusing classifier outputs. TREC 2004 dataset is used containing frames such as boat, beach, basket and airplane from news and advertisement which are not semantic scenes categories. Another classification system is proposed in [8] where individual classifiers are trained on color and texture and they are fused using a third SVM classifier which is called as high-level concept classifier. The dataset contains sixty categories all of which are objects and animals. There is no scene categories included and the proposed classifier fusion scheme is not compared with any other fusion approach. A very similar work on indoor/outdoor classification is described in [9] where the authors use two individual SVM classifiers based on color and texture features. Another SVM classifier is trained on the output of these two classifiers to perform a binary indoor/outdoor classification.

In spite of all these works, a systematical and comparative study of feature level fusion and classifier fusion techniques for image classification is still missing. Proposed methods in the literature are generally not compared with other methods. Even in a very complete comparison of audio data classification techniques [10], where weighted and unweighted voting strategies are discussed, the feature level fusion approach is missing. The datasets used in these studies contain very specific categories like rock images. However, existing work show that the nature of the data influences classification performance.

In this paper we compare these two approaches for natural scenes using a variety of visual descriptors. We evaluate our results with commonly used performance criteria. The rest of the paper is organized as follows: Section 2 introduces pre-classification and post classification techniques. In section 3 features used for image representation are briefly presented. Measures used to evaluate classification performances are described in section 4. Experimental results are given in section 5 and finally conclusion in section 6.

## II. FEATURE LEVEL FUSION

Feature level fusion refers to combining information prior to the application of any classifier or matching algorithm [11]. It is performed by concatenating individual vectors to form a single feature vector. Concatenation of vectors can lead to two major problems. Firstly, the resulting feature vector may have a very large dimensionality, a problem referred to as the 'curse of dimensionality'. There exist a number of techniques to remedy the curse of dimensionality problem by reducing the dimensionality of a feature vector. The most frequently used one is Principal Component Analysis (PCA).

The second problem is the scale effect due to different magnitude of numerical values of the individual feature vectors. Scale effects can be addressed by re-scaling or normalizing feature vectors. A detailed study of normalization techniques can be found in [11].

## III. CLASSIFIER FUSION

Classifier fusion refers to combining information after the classification or matching stage. Information to combine is the classifier output which can either be a class label or numerical output value.

### A. Class Label Fusion

#### 1) Majority Voting

Classifiers that produce class labels are generally fused using a voting method. A generalized voting definition is given below.

The decision vector d formed by the output of the classifiers is defined as: $d = |d_1, d_2, ..., d_n|^T$ where $d_i \in {c_1, c_2, ... c_m}$, $c_i$ denotes the label of the $i^{th}$ class. Let binary characteristic function be defined as follows:

$$B_j(c_i) = \begin{cases} 1 & if \ d_j = c_i \\ 0 & if \ d_j \neq c_i \end{cases} \quad (1)$$

Then the general voting routine can be defined as:

$$E(d) = \begin{cases} c_i & if \ \forall_{t \in \{1,...,m\}} \sum_{j=1}^{n} B_j(c_t) \leq \sum_{j=1}^{n} B_j(c_i) \geq \alpha \cdot m \\ r & otherwise \end{cases} \quad (2)$$

The case where $\alpha = 0.5$ is commonly known as the majority vote.

### B. Soft Output Fusion

#### 1) Bayes Average Method

The Bayesian methods can be applied to the classifier fusion under the condition that the outputs of the classifier are expressed in posterior probabilities. Effectively combination of given likelihoods is also a probability of the same type, which is expected to be higher than the probability of the best individual classifier for the correct class.

If the outputs of the multiple classifier system are given as posterior probabilities that an input sample x comes from a particular class $C_i$, I=1,..,m: $P(x \in c_i/x)$, it is possible to calculate an average posterior probability taken from all classifiers:

$$P_E(x \in c_i / x) = \frac{1}{K} \sum_{k=1}^{K} P_K(x \in C_i / x) \quad (3)$$

#### 2) Bayes Belief Integration

The approach mentioned above treats equally all the classifiers and does not explicitly consider different errors produced by each of them. These errors can be comprehensively described by means of confusion matrix given by:

$$PT_k = \begin{pmatrix} n_{11}^{(k)} ... n_{1(M+1)}^{(k)} \\ ... \\ n_{M1}^{(k)} ... n_{M(M+1)}^{(k)} \end{pmatrix} \quad (4)$$

where rows correspond to classes: $c_1,...,c_M$ from which the input sample was drawn from and columns denote the classes to which the input sample was assigned by the classifier $e_k$. The values $n_{ij}^{(k)}$ express how many input samples coming from class $c_i$ were assigned to class $c_j$. On the basis of the

confusion matrix $PT_k$ it is possible to build the belief measure of correct assignment as given by:

$$Bel(x \in c_i / e_k(x)) = P(x \in c_i / e_k(x) = j_k) \quad (5)$$

Where i=1,...,M; j=1,...,M+1 and

$$P(x \in c_i / e_k(x) = j) = \frac{n_{ij}^k}{\sum_{i=1}^{M} n_{ij}^{(k)}} \quad (6)$$

Having defined such a belief measure for each classifier we can combine them in order to create new belief measure of the multiple classifier system as follows:

$$Bel(i) = P(x \in c_i) \frac{\prod_{k=1}^{K} P(x \in c_i / e_k(x) = j_k)}{\prod_{k=1}^{K} P(x \in c_i)} \quad (7)$$

The probabilities used in the above formula can be easily estimated from the confusion matrix. The class with the highest combined belief measure: Bel(I) is chosen as a final classification decision. Alternatively selection of any class may be rejected if the combined belief is smaller than a specified threshold value. The decision function for a new instance x is as follows:

$$c(x) = \arg\max_{1 \leq i \leq K} Bel(i) \quad (8)$$

## IV. EXPERIMENTAL SETUP

### A. Visual Descriptors

Existing image categorization systems in the literature can be generally classified into two categories based on the underlying framework for image content representation. In the first category the image is segmented into some meaningful components that are used as semantic elements to characterize image content. Or a regular grid is applied to divide the image in sub blocks. The second category takes an image as a whole visual appearance and characterizes image contents by using image-based global visual features. In a previous work we evaluated multi-class classification strategies for SVM. We have compared different image representations and concluded that texture leads to the highest classification accuracy as a local representation, while gist is the most performing global representation. We used feature level fusion to combine these representations for multi-class classification [12].

In this work we use two extra representations which are color layout descriptor and edge orientation histogram. Color layout descriptor is a compact and resolution-invariant MPEG-7 visual descriptor defined in the YCbCr color space and designed to capture the spatial distribution of color in an image or an arbitrary-shaped region. The feature extraction process consists of four stages. The input image is partitioned into 8x8 = 64 blocks, each represented by its average color. A DCT transformation is applied on the resulting 8x8 image. The resulting coefficients are zig-zag-scanned and only 6 coefficients for luminance and 3 for each chrominance are kept, leading to a 12-dimensional vector. Finally, the remaining coefficients are quantized [13].

Edge histogram descriptor captures the spatial distribution of edges. Four directions of edges (0 ±, 45 ±, 90 ±, 135 ±) are detected in addition to non-directional ones. The input image is divided in 16 non-overlapping blocks and a block-based extraction scheme is applied to extract the five types of edges and calculate their relative populations, resulting in a 80-dimensional vector.

Our texture representation is obtained by extracting four attributes namely energy, entropy, homogeneity and inertia from gray level co-occurrence matrix. This feature is extracted from blocks of 64x64 pixels. Since our images are 256x256 pixel it leads to 16x4=64-dimensional vector. We use gist to characterize spectral information. Gist is a low dimensional representation of the scene structure based on the output of filters tuned to different orientations and scales[14]. Our gist feature is a 476-dimensional vector.

### B. Image Database

Our image database contains 8 categories of natural scenes: highway(260), streets(292), forest(328), open country(410), inside of cities(308), tall buildings(356), coast(360) and mountain(374) images (Numbers in brackets represent the size of each categories). The database provided by Oliva and Torralba was collected from a mixture of COREL images as well as personal photographs [14]. All images are colored and sized of 256x256 pixels. For each classification experiment 100 images of each category are reserved for test purpose and the remaining images are used as training set. Samples images for the 8 categories are given in Figure 1.

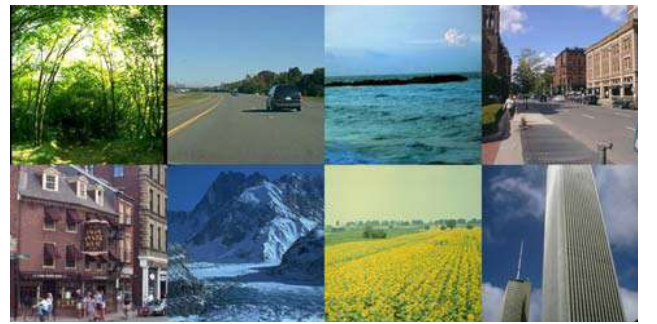

**Fig. 1.** Sample images of the database. From top left to bottom right: Forest, Highway, Coast, Street, Inside of city, Street, Mountain, Open Country, Tall building.

We used our image database to obtain 4 couples such that they are visually the most similar categories. These couples are

Street-Inside of city, Tall building-Street, Mountain -Tall building and Mountain-Inside of city. The reason why we have chosen the most similar categories is to obtain worst case scenarios for binary classification. We suppose that if two classes are similar then the binary classification performance for these classes is low and vice versa. In other words, similarity of two classes varies in the opposite way with binary classification accuracy of these classes. The strategy used to find the most similar classes is described in a previous work [12]. It is in accordance with visual judgment. Note that the street scene and inside of city scene in figure 2 are very similar but they belong to different classes.

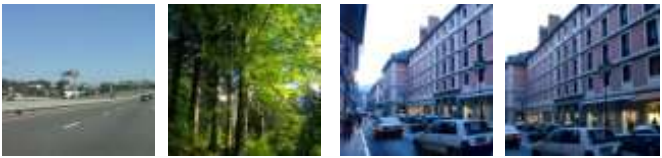

**Fig. 2** Sample images of two least similar and two most similar classes. From left to right: Highway, Forest, Street and Inside of city.

## V. EXPERIMENTAL RESULTS

### A. Feature Level Fusion

To evaluate feature fusion strategies, we performed a set of binary classification experiments using the most similar classes. The reason for choosing the most similar classes is to see the difference between results in a larger scale. We preferred binary classification for the simplicity and clarity of the results. It is sufficient to illustrate the difference between the compared strategies.

In each experiment we used all of the image representation described in paragraph 3. For feature level fusion these representations are fused by concatenating the corresponding feature vectors. Normalization can be performed in two ways. The first one is to normalize each individual feature vector then concatenate those in order to obtain one large vector. We term this as pre-normalization. The second way is to concatenate the individual feature vectors first then to normalize the obtained larger vector. This can be termed as post-normalization. PCA is applied like the post-normalization. A large vector that contains multiple feature vectors is processed to obtain the principle components. The resulting feature vector is smaller in dimension.

In a previous work we had compared several normalization functions and we concluded that z-score lead to the highest accuracies among the other functions namely decimal, minmax, tanh and median normalization functions. Therefore we used the z-score as a normalization function in our experiments to transform the features into a common numerical range.

Classification results using feature level fusion for the four modalities in terms of accuracy are given in the following table.

TABLE I. FEATURE LEVEL FUSION

| Modalities | Pre Normalization | Post Normalization | PCA |
|---|---|---|---|
| Street-Inside of city | 0.55 | 0.53 | 0.58 |
| Tall building-Street | 0.69 | 0.66 | 0.71 |
| Mountain -Tall building | 0.66 | 0.69 | 0.70 |
| Mountain-Inside of city | 0.74 | 0.75 | 0.77 |

According to these results PCA is more performing than normalization techniques independently of the modalities which is in accordance with the results in the literature. Pre-normalization and post-normalization lead to very similar results however pre-normalization is slightly better. This can be explained by the original input distribution of the features is better retained by performing the pre-normalization.

This conclusion indicates that it is preferable to use PCA not only for the better classification accuracy but also for the performance increment in training phase due to the smaller size of feature vector.

### B. Classifier Fusion

TABLE II. CLASSIFIER FUSION

| Modalities | Majority Voting | Bayes Average | Bayes Belief Integration |
|---|---|---|---|
| Street-Inside of city | 0.62 | 0.66 | 0.66 |
| Tall building-Street | 0.75 | 0.78 | 0.80 |
| Mountain - Tall building | 0.77 | 0.81 | 0.83 |
| Mountain-Inside of city | 0.83 | 0.88 | 0.88 |

We compared majority voting as class label fusion, Bayes average and Bayes Belief Integration as soft output fusion methods using the same modalities. Classification results are shown in Table 2. The most performing method is Bayes belief integration with a slight difference with Bayes average method.

One should note that soft output fusion strategies lead to better performance than class label fusion. This means that fusing numerical classifier outputs is preferable than fusing decisions. This is due to better interpretation of the information provided by the classifiers.

## C. Comparison of Feature Level Fusion and Classifier Fusion

Table 3 illustrates the classification results obtained using the best methods derived from both strategies. Results show that Bayes Belief Integration method outperforms PCA in each of the binary classifications. This result shows that classifier fusion is advantageous for this task.

TABLE III. FEATURE LEVEL FUSION VS. CLASSIFIER FUSION

|  | *Feature Level Fusion* | *ClassifierFusion* |
|---|---|---|
| *Modalities* | PCA | BayesBeliefIntegration |
| Street-Inside of city | *0.58* | *0.66* |
| Tall building-Street | *0.71* | *0.80* |
| Mountain -Tall building | *0.70* | *0.83* |
| Mountain-Inside of city | *0.77* | *0.88* |

For each of the binary classifications, accuracies are higher in classifier fusion. This can be explained by the fact that the classifier fusion methods use more resources then feature fusion methods. For instance, to perform a binary classification using classifier fusion, four individual binary classifiers are trained each of which on a single feature. On the other hand using feature level fusion this task could be carried out using only one binary classifier.

Classifier fusion outperforms feature level fusion. The least performing classifier fusion method which is the majority voting leads to better results than the most performing feature level fusion.

## VI. CONCLUSION

We have presented a comparison feature level fusion and classifier fusion for natural scene image classification using SVM. According to our results classifier fusion strategies seem to perform better that the feature level fusion strategies. This conclusion is confirmed with all experiments performed on four modalities of image groups using all types of image representation namely color, texture, edge and gist.

In order to improve classification accuracies one should treat image representation individually not concatenate them. Which means classifier fusion is more appropriate then feature level fusion to have high classification performance. Furthermore, fusion of individual classifiers should be done in a numerical output level not in decision level.